\documentclass[letterpaper, 10 pt, conference]{ieeeconf}  

\IEEEoverridecommandlockouts                              

\usepackage{booktabs}   
\usepackage{graphicx}   
\usepackage{tabularx}
\usepackage{amsmath}
\usepackage{amssymb}
\usepackage{dblfloatfix}  
\usepackage{float}

\usepackage{hyperref}
\hypersetup{
   colorlinks=true,
   linkcolor=blue,
    filecolor=magenta,      
   urlcolor=cyan,
}

\usepackage{array}
\newcolumntype{Y}{>{\centering\arraybackslash}X}

\usepackage{booktabs, tabularx, multirow}

\usepackage{caption}
\usepackage{graphicx}
\usepackage[justification=raggedright,singlelinecheck=false]{caption}

\makeatletter
\long\def\@makecaption#1#2{%
  \vskip\abovecaptionskip
  \small #1. #2\par
  \vskip\belowcaptionskip
}
\makeatother

\begin{document}

\title{\LARGE \bf
DexViTac: Collecting Human Visuo-Tactile-Kinematic Demonstrations for Contact-Rich Dexterous Manipulation
}

\author{
    \textbf{Xitong Chen$^{1}$,  Yifeng Pan$^{1}$,  Min Li$^{1*}$ and Xiaotian Ding$^{2}$} \\
    $^{1}$\textnormal{State Key Laboratory of Intelligent Manufacturing Equipment and Technology,} \\
    \textnormal{Huazhong University of Science and Technology, Wuhan, China} \\
    $^{2}$\textnormal{Wuhan Huaweike Intelligent Technology Co., Ltd., Wuhan, China} \\
    \noalign{\vskip 1ex}
}



\makeatletter
\def\IEEEaftertitletext#1{\gdef\@IEEEaftertitletext{#1}}
\makeatother
\IEEEaftertitletext{%
  \begin{center}
  \vspace{-0.4em}
  \captionsetup{type=figure}
  \includegraphics[width=\textwidth,height=0.32\textheight,keepaspectratio]{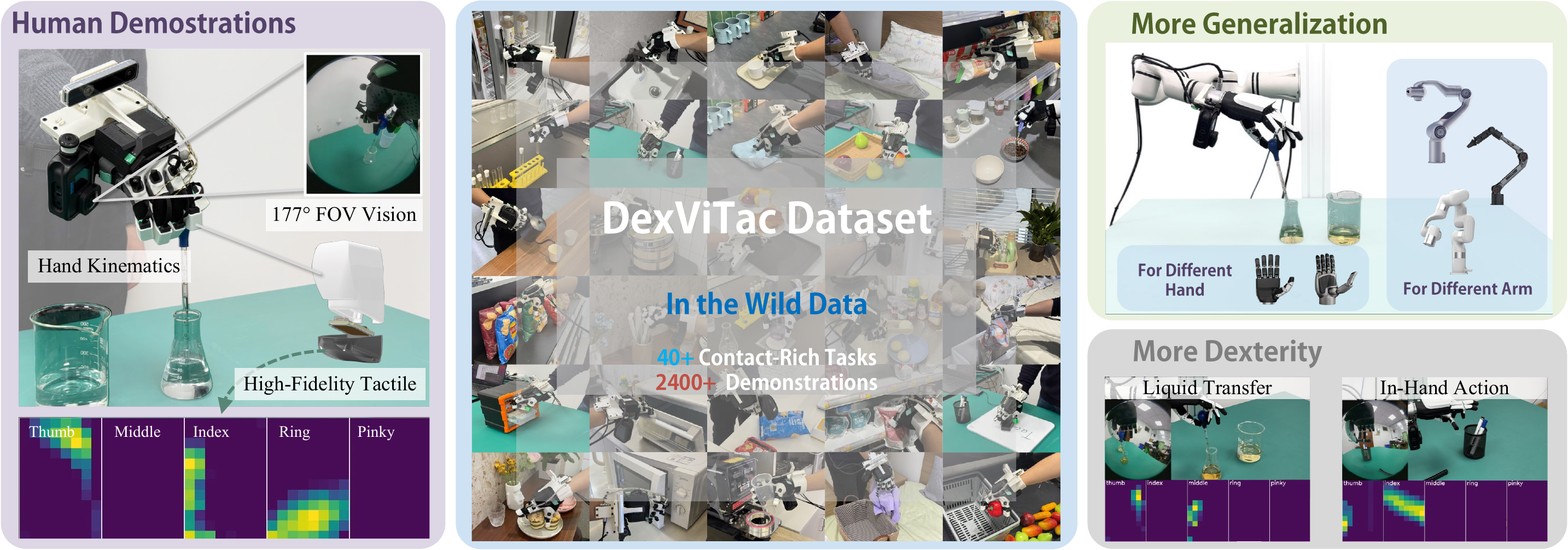}
  \captionof{figure}{\textbf{Overview of DexViTac.} DexViTac is a portable, human-centric data collection system designed for contact-rich dexterous manipulation. It enables the high-fidelity acquisition of first-person vision, high-density tactile sensing, and hand kinematics within unstructured, in-the-wild environments, facilitating the development of generalized robotic policies across diverse hardware platforms.}
  \label{fig:overview}
  \vspace{-0.4em}
  \end{center}
}


\maketitle
\thispagestyle{empty}
\pagestyle{empty}

\begin{abstract}

Large-scale, high-quality multimodal demonstrations are essential for robot learning of contact-rich dexterous manipulation. While human-centric data collection systems lower the barrier to scaling, they struggle to capture the tactile information during physical interactions. Motivated by this, we present DexViTac, a portable, human-centric data collection system tailored for contact-rich dexterous manipulation. The system enables the high-fidelity acquisition of first-person vision, high-density tactile sensing, end-effector poses, and hand kinematics within unstructured, in-the-wild environments. Building upon this hardware, we propose a kinematics-grounded tactile representation learning algorithm that effectively resolves semantic ambiguities within tactile signals. Leveraging the efficiency of DexViTac, we construct a multimodal dataset comprising over 2,400 visuo-tactile-kinematic demonstrations. Experiments demonstrate that DexViTac achieves a collection efficiency exceeding 248 demonstrations per hour and remains robust against complex visual occlusions. Real-world deployment confirms that policies trained with the proposed dataset and learning strategy achieve an average success rate exceeding 85\% across four challenging tasks. This performance significantly outperforms baseline methods, thereby validating the substantial improvement the system provides for learning contact-rich dexterous manipulation. Project page: \url{https://xitong-c.github.io/DexViTac/}.

\end{abstract}

\section{INTRODUCTION}

Realizing general-purpose dexterous manipulation is a core objective in robotics research. Inspired by scaling laws, the development of robust and generalizable policies relies on the acquisition of large-scale data. In complex physical interactions, the tactile modality provides critical contact information that is often absent in visual observations, serving as a cornerstone for stable interaction control \cite{Effecttactile}.

However, the acquisition of high-quality tactile data from human demonstrations remains a significant challenge. Traditional teleoperation \cite{mobilealoha}, \cite{Dexpilot}, as well as Augmented Reality (AR) and Virtual Reality (VR) systems \cite{bunny}, are typically restricted to structured environments and lack multi-finger tactile feedback. Embodiment-agnostic frameworks such as UMI \cite{UMI} are primarily limited to low-degree-of-freedom grippers(DOF), while existing exoskeleton \cite{Exo-ViHa}, \cite{DOglove} solutions tend to be intrusive to natural motion or lack essential contact information. 

The effective extraction and utilization of high-dimensional tactile data for policy learning presents another major hurdle. High-fidelity tactile signals exhibit inherent spatial semantic ambiguity, where local signals lack physical meaning when dissociated from the global kinematic configuration of the hand \cite{tacqiyi}. Consequently, there is an urgent need for systems capable of acquiring visuo-tactile-kinematic demonstration pairs at scale in unstructured environments, coupled with integrated representation learning methods.

To address these challenges, we introduce DexViTac, a human-centric data collection system tailored for contact-rich dexterous manipulation. By integrating fish-eye camera, motion-capture gloves, high-resolution tactile sensors, and a T265 camera, the system facilitates the acquisition of tightly coupled visuo-tactile-kinematic demonstrations without impeding natural human motion. To resolve the semantic ambiguity of tactile signals, we propose a kinematics-grounded tactile representation learning framework. Leveraging this representation learning method, we further develop a two-stage training strategy. Real-world robotic experiments demonstrate that the system achieves an average success rate exceeding 85\% across four challenging, contact-rich manipulation tasks, thereby validating the efficacy of the data collected by DexViTac. 

To foster further research in contact-rich manipulation, we will open-source the complete hardware designs, the large-scale dataset, and the associated codebases. In summary, our main contributions are:
\begin{itemize}
\item  \textbf{DexViTac Data Collection System:}A portable, human-centric system for contact-rich dexterous manipulation. By integrating first-person vision, high-density tactile sensing, end-effector poses, and hand kinematics, it acquires high-fidelity human demonstrations in unstructured, in-the-wild environments.
\item  \textbf{Kinematics-Grounded Tactile Representation:} A self-supervised tactile-kinematic learning framework that resolves semantic ambiguity by grounding local tactile features within a global kinematic latent space, ensuring physically consistent multi-finger coordination. 
\item \textbf{Large-scale contact-rich dataset:} A dataset spanning 10$+$ environments and 40$+$ tasks, comprising 2,400$+$ visuo-tactile-kinematic demonstrations .
\end{itemize}

\section{RELATED WORK}
\subsection{Scalable Data Collection for Dexterous Manipulation}

Large-scale data collection serves as a cornerstone for robotic generalization. Human-centric and  Embodiment-agnostic frameworks have emerged as a dominant paradigm: UMI \cite{UMI} utilizes handheld grippers and fisheye cameras to achieve high-efficiency data collection in the wild. However, such systems are structurally constrained and cannot support multi-finger dexterous manipulation. Recent studies, including DexCap \cite{dexcap}, DexWild \cite{dexwild}, and DexUMI \cite{xu2025dexumi}, incorporate kinematic information to enhance demonstration consistency, while others, such as BunnyVisionPro \cite{bunny}, Open-TeleVision \cite{cheng2024opentelevision} and ARCap \cite{arcap}, leverage VR/AR to improve interaction naturalness. Nevertheless, these methods focus predominantly on vision and proprioception, often overlooking the tactile modality, which is essential for physical interaction. Current portable frameworks lack high-fidelity contact data, thereby limiting the potential for fine-grained force regulation. Consequently, an urgent need exists to develop human-centric data collection systems that integrate high-fidelity tactile sensing for five-finger dexterous hands .

\subsection{Visuo-Tactile Learning for Contact-Rich Manipulation}

Contact-rich manipulation requires the effective learning of high-dimensional visual representations and local tactile physical constraints. The core challenges involve overcoming modal collapse and the sim-to-real distribution shift. Existing research enhances consistency through self-supervised exploration \cite{DexterityfromTouch}, masked reconstruction \cite{touchinthewild}, contrastive learning \cite{vitamin}, and unified sensor tokens \cite{Binding}. To improve robustness, some methods utilize sparse or binary signals to decouple tasks \cite{VTDexManip}, \cite{10.1126}.However, such techniques often result in the loss of high-frequency information and limit fine-grained force regulation.  

Real-world deployment is hindered by cross-modal spatio-temporal misalignment and physical embodiment differences. Regarding spatial mapping, existing methods introduce 3D tactile normalization \cite{Canonical} or LLM-based contact mode inference \cite{DextER}. For temporal latency, techniques such as future tactile prediction \cite{ViTacFormer} or optical flow hallucination \cite{FBI} are used to suppress jitter. However, these "patches" do not address the underlying spatio-temporal heterogeneity and overlook the tactile semantic ambiguity inherent in multi-finger interactions. To this end, we propose a hardware-software synergistic paradigm: introducing a kinematics-grounded framework that strongly constrains local tactile data within the hand kinematics space. This design effectively mitigates semantic ambiguity and provides physically consistent, robust cross-modal representations, bridging the human-robot embodiment gap.

\section{METHOD}

\begin{figure*}[h]
\centering
\includegraphics[width=\textwidth]{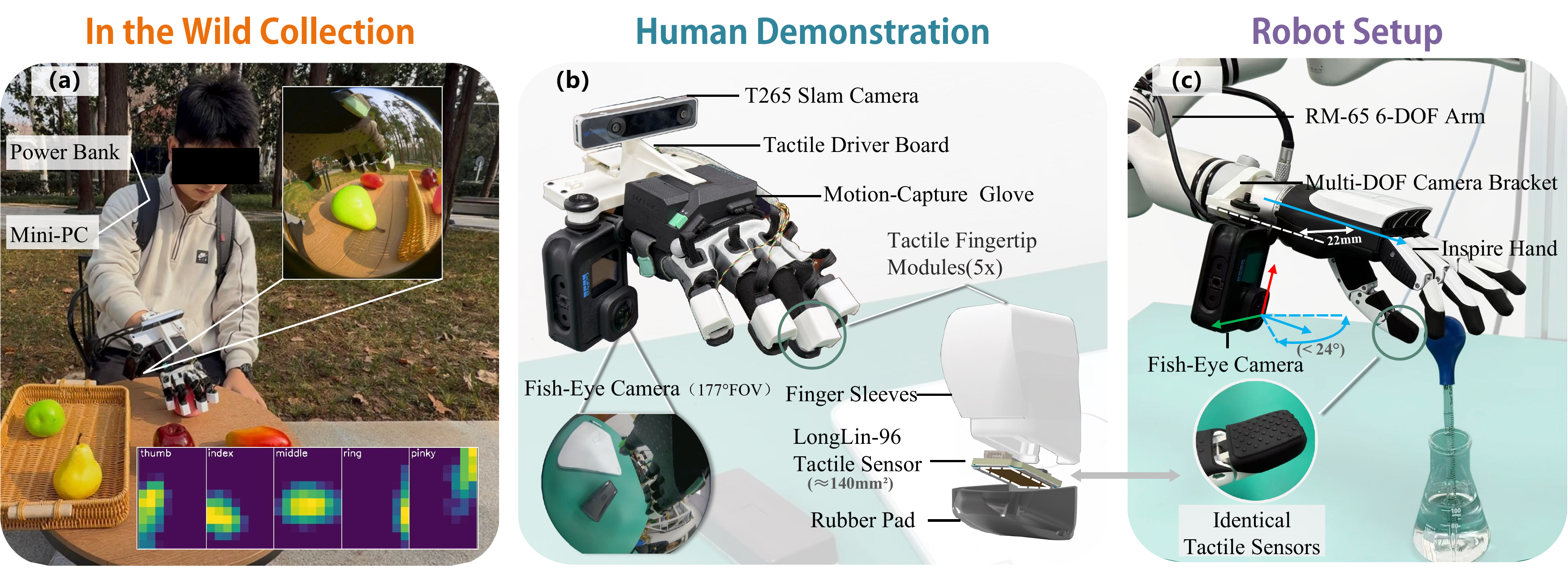}
\caption{\textbf{Hardware design.} (a) Equipped with a backpack-integrated Mini-PC and power bank, the proposed system enables out-of-the-box multimodal data collection within in-the-wild environments. (b) The human demonstration interface features a decoupled design comprising a fisheye camera, motion-capture gloves, high-resolution tactile sensors, and a T265 tracking camera. (c) The robot execution platform utilizes an isomorphic perception architecture wherein the tactile sensors remain strictly consistent with those on the human demonstration interface.}
\label{fig:hardware}
\end{figure*}

DexViTac presents a high-fidelity, human-centric universal data collection solution for contact-rich dexterous manipulation. This section first elaborates on the symmetric hardware-software synergistic architecture (Sec. III-A, B); next, it discusses the data preprocessing methods designed to address human-robot embodiment heterogeneity (Sec. III-C); finally, the effectiveness of the collected data in unstructured, contact-rich tasks is validated through a two-stage learning strategy (Sec. III-D)

\subsection{Hardware Design}

The hardware design of DexViTac follows three principles:

\begin{itemize}
\item  \textbf{Tactile-Centric:} Capture complex contact information in human demonstrations with high fidelity.
\item  \textbf{Anthropomorphic Dexterity:} Precisely track five-finger articulation during complex tasks. 
\item \textbf{Embodiment-Agnostic:} Utilize a robot-free design to support in-the-wild data collection and cross-hardware generalization for large-scale, scalable acquisition.
\end{itemize}

The complete system comprises a human demonstration interface and a robotic execution platform. These two components feature a mirrored sensor configuration and identical data modalities to minimize the human-robot domain gap and enhance the transferability of policy learning. 
 
\textbf{Human Demonstration.} This component is designed to capture high-dimensional asymmetric multimodal information synchronously without interference. As shown in Fig. 2 (b), the system employs a decoupled design to ensure acquisition flexibility, data integrity, and scalability, consisting primarily of the following four modules:

\textbf{1) Tactile Sensing Module:} The system integrates customized tactile modules with motion-capture gloves, featuring high-density HIT LongLin-96 fingertip tactile sensors. These flexible pressure-sensitive sensors provide an $8 \times 16$ taxel array (140 ${mm}^2$ area) with a 0--20 $N$ range and 0.01 $N$ sensitivity. Operating at a sampling rate of  1000Hz an accuracy of $<$ 5\% full-scale (FS), they capture millisecond-level pressure distributions and textures. Notably, identical sensors are deployed on robotic end-effectors, eliminating hardware-level perceptual domain gaps. 

\textbf{2) Hand Kinematics Capture: }We employ Manus Quantum Metagloves equipped with Inertial Measurement Units (IMUs)  for hand pose tracking. Compared to vision-based methods, this approach is robust against occlusions and close-contact scenarios, ensuring high-fidelity acquisition of fine-grained finger states.

\textbf{3) First-Person Wide-Angle Vision:} The visual module utilizes a GoPro Hero 12 Black with a Max Lens Mod 2.0 fisheye lens, offering a 177$^{\circ}$ Field of View (FOV) for comprehensive coverage of near-field hand-object interactions and environmental context. This wide-angle configuration provides rich visual information while enhancing system portability. Furthermore, the inherent fisheye distortion yields higher effective spatial resolution near the image center, improving the visual representation of manipulated objects. 

\textbf{4) Global Pose Tracking:} We utilize the built-in Visual-Inertial Odometry (VIO) algorithm of a RealSense T265 camera to output 6-DOF poses and confidence levels without requiring external calibration. Combined with a custom error-compensation algorithm, this setup enables stable and precise tracking of wrist trajectories.

\textbf{Robot Setup.} Fig.~2(c) illustrates the isomorphic sensing architecture, which supports multi-platform compatibility through modular interfaces. In our experiments, robot end-effector poses and tactile array data from the dexterous hand are directly accessed without the need for additional sensors. To ensure the first-person perspective remains consistent with human demonstrations, the GoPro camera is mounted below the wrist using a multi-DOF bracket. The camera orientation is adjusted such that the optical center aligns with the middle fingertip of the dexterous hand, maintaining a horizontal deviation within $\pm $12$^{\circ}$.

\subsection{Data Collection Software Framework}

Building upon the DexViTac hardware, we develop an integrated data collection framework as depicted in Fig.~3. This architecture enables the efficient acquisition, spatiotemporal alignment, and formatted storage of heterogeneous and asynchronous multimodal data.

\begin{figure}[h]
\centering
\includegraphics[width=1\linewidth]{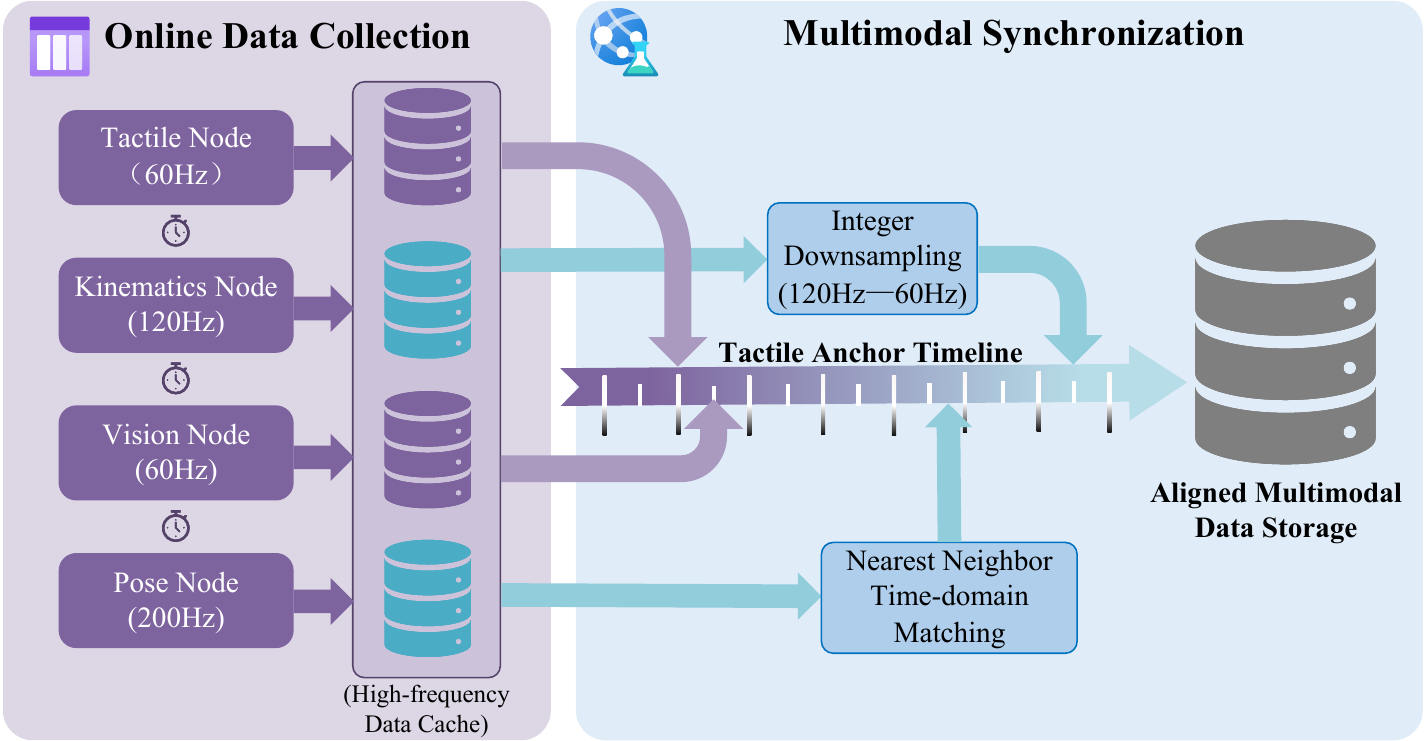}
\caption{\textbf{Data collection pipeline.} To prevent frame loss and ensure tight spatiotemporal alignment across different modalities, we employ high-frequency buffering alongside a tactile-anchored synchronization strategy that involves downsampling and nearest-neighbor matching.}
\label{fig:workflow}
\end{figure}

\textbf{Global Data Collection.} The  data collection pipeline is built upon ROS2, with tactile, vision, kinematics, and pose-tracking modules operating as independent nodes. A central node executes timestamp alignment, frequency coordination, and buffer management. The tactile node samples high-resolution distributions from all five fingers, with each frame structured as five 2D matrices to provide precise force sensing. To account for varying hand morphologies, the kinematics node moves away from absolute joint angles, instead employing $Stretch$ (flexion-extension) and $Spread$ (abduction-adduction) parameters to express 19\text{-DOF} motion. This representation characterizes only relative states and topological relationships, balancing precision with cross-operator generalization and laying the foundation for retargeting. The vision and pose nodes capture GoPro images (1920$\times$1080, 60Hz) and T265 real-time 6\text{-D} poses, respectively. The lightweight, modular design eliminates the need for external SLAM or complex calibration. Combined with a portable computing unit, the system is plug-and-play in unstructured, in the wild environments (Fig. 2(a)), significantly lowering the barrier to large-scale data collection.

\textbf{Tactile-Dominant Multimodal Time Synchronization.} In asynchronous sensor systems, precise alignment is critical for policy learning. DexViTac employs a two-stage "hardware buffering-software alignment" synchronization mechanism. During online collection, nodes share a ROS2 system clock and utilize thread-safe buffer queues to prevent frame loss. In the offline stage, we propose a tactile-led semantic alignment strategy to handle heterogeneous sampling rates. Given the high temporal sensitivity of tactile events (e.g., contact, release), the algorithm sets tactile (60Hz) timestamps as global anchors to preserve key semantic information. Based on these anchors, vision streams (60Hz) undergo hardware frame-level alignment; hand kinematics (120Hz) use integer downsampling; and high-frequency global poses (200Hz) employ binary search for nearest-neighbor matching. This strategy ensures the causal integrity of contact states, generating strictly synchronized observation-action pairs.

\subsection{Data Preprocessing}

As a human-centric system, the raw data of DexViTac exhibits a significant human-robot domain gap. Three core issues must be resolved before feeding the data into the policy network: \textbf{(1) Kinematic Adaptation:} How to map high-dimensional human hand configurations to heterogeneous dexterous hands while overcoming DOF-mismatch and maintaining operational semantic consistency; \textbf{(2) Spatial Alignment:} How to retarget T265 local poses to the global coordinate system of the robot to achieve cross-embodiment pose generalization; and \textbf{(3) Tactile Fidelity:} How to suppress sensor noise and recover contact sparsity to prevent spurious signals from undermining the policy's causal reasoning regarding physical interactions?

\textbf{Data Retargeting:} Given the embodiment-agnostic nature of DexViTac, we establish a two-layer mapping mechanism from the human demonstration domain $D_H$ to the robot execution domain $D_R$.

\textbf{1) Hand Retargeting:} In human demonstrations, the state at time $t$ captured by motion-capture gloves is denoted as $h_t \in \mathbb{R}^{D_H}$ (based on $stretch/spread$ parameters), and the robot joint command is denoted as $q_t \in \mathbb{R}^{D_R}$. Due to differences in hardware structures, typically $D_H \neq D_R$. To eliminate embodiment differences, we introduce a retargeting mechanism during the preprocessing stage:
$$q_t = \mathcal{F}_R\big( \mathcal{F}_H(h_t) \big) \eqno{(1)}$$
where $\mathcal{F}_H$ maps human hand parameters to the kinematic joint space defined by the target actuator's URDF, and $\mathcal{F}_R$ further maps it to the robot control space (which can be solved using PyBullet inverse kinematics tools). By employing a decoupled intermediate representation, while keeping $h_t$ and $\mathcal{F}_H$ constant, one only needs to reset $\mathcal{F}_R$ to adapt to different dexterous hands, achieving generalized reuse across various hardware configurations.

\textbf{2) Robot Arm Retargeting:} At time $t$, the T265 outputs its motion $(\mathbf{p}_{\mathrm{cam}}^{(t)}, \mathbf{R}_{\mathrm{cam}}^{(t)})$ relative to the initial pose in its local coordinate frame $\{\mathrm{cam}\}$. To map this trajectory to the robot end-effector, we employ an incremental retargeting algorithm, where the end-effector pose in the base frame $\{\mathrm{base}\}$ is defined as:

$$\mathbf{p}_{\mathrm{ee}}^{(t)} = \mathbf{p}_{\mathrm{ee}}^{(0)} + \mathbf{R}_{\mathrm{c2b}} \mathbf{p}_{\mathrm{cam}}^{(t)} \eqno{(2)}$$
$$\mathbf{R}_{\mathrm{ee}}^{(t)} = \mathbf{R}_{\mathrm{c2b}} \mathbf{R}_{\mathrm{cam}}^{(t)} \mathbf{R}_{\mathrm{ee}}^{(0)} \eqno{(3)}$$

where $(\mathbf{p}_{\mathrm{ee}}^{(0)}, \mathbf{R}_{\mathrm{ee}}^{(0)})$ represents the initial pose of the robot arm. The rotation matrix $\mathbf{R}_{\mathrm{c2b}}$ is obtained through an automatic calibration procedure: the user moves the $T265$ along the axes of the base frame as prompted, allowing the program to record and solve for the transformation. In practical deployment, only $\mathbf{R}_{\mathrm{c2b}}$ needs to be re-solved to adapt the demonstration data to different robot arm platforms, achieving cross-configuration generalization.

\textbf{3) Tactile Denoising:} Raw tactile arrays are susceptible to sensor hysteresis and thermal drift, resulting in spatially non-uniform zero-point noise (often referred to as ghost-touches) even in the absence of contact. To eliminate false-positive interference and enhance contact sparsity, we implement a soft-thresholding process based on dynamic baselines. Let the raw tactile array at frame $t$ be $T_t^{\mathrm{raw}} \in \mathbb{R}^{H \times W}$ and the no-load bias be $T^{\mathrm{bias}}$. The denoised representation $T_t^{\mathrm{clean}}$ is calculated as follows:
$$T_t^{\mathrm{clean}} = \max\left( 0,\, T_t^{\mathrm{raw}} - T^{\mathrm{bias}} - \epsilon \right) \eqno{(4)}$$
where $\epsilon$ denotes the noise floor tolerance. This operator effectively zeros out background noise while preserving subtle contact signals, preventing invalid data from interfering with the causal reasoning of the policy network.

\subsection{Pretraining and Policy Learning}

\begin{figure*}[!htbp]
\centering
\includegraphics[width=\textwidth]{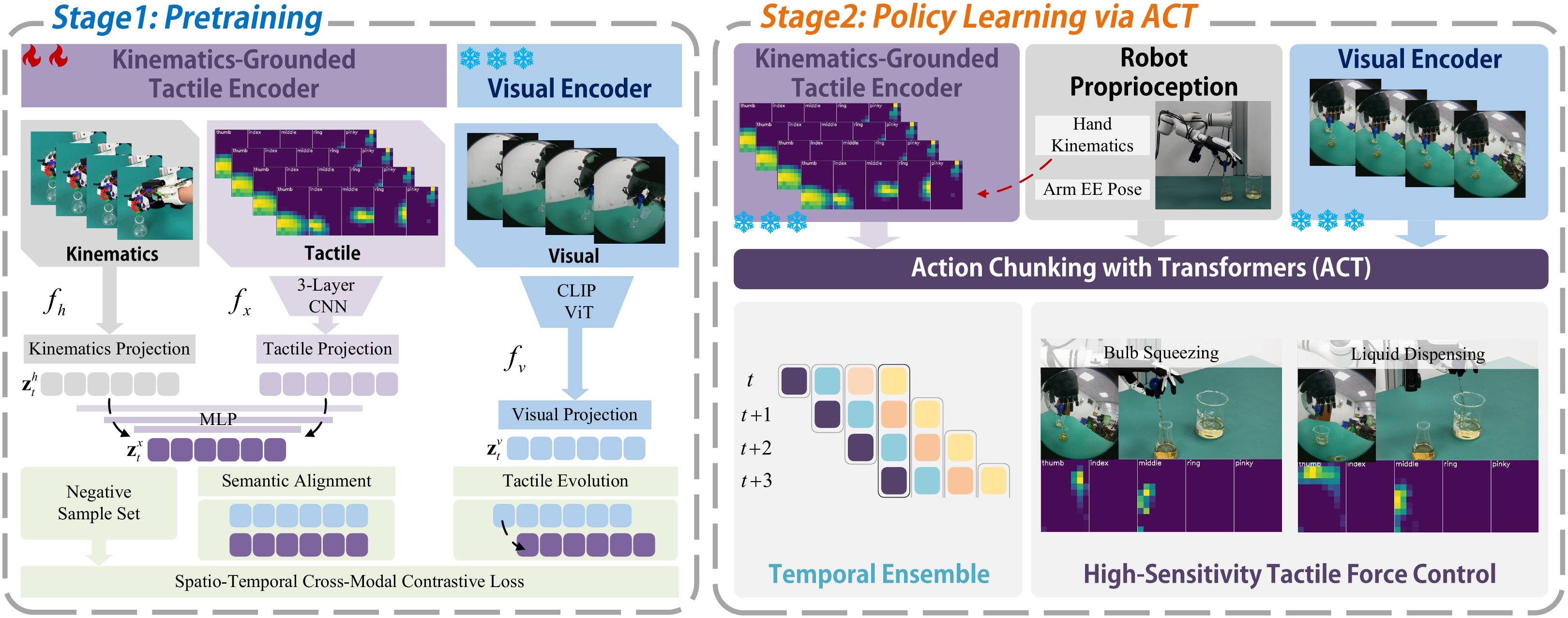}
\caption{\textbf{Two-stage learning strategy.} Stage $1$: A self-supervised framework aligns high-density tactile features with visual anchors utilizing a kinematics-Grounded encoder to learn spatially anchored representations. Stage $2$: The pretrained encoders are subsequently integrated into an Action Chunking with Transformers (ACT) policy to map synchronized multimodal observations to multi-step action sequences for contact-rich dexterous manipulation.}
\label{fig:pretrain}
\end{figure*}

We adopt a two-stage training strategy (Fig. 4): first, we learn kinematics-conditioned and temporally consistent visuo-tactile representations through self-supervised pre-training; subsequently, these representations are transferred to the Action Chunking with Transformers (ACT) policy to achieve efficient learning of dexterous manipulation.

\textbf{Kinematics-Grounded Tactile Pretraining.} This stage aims to learn an encoder capable of simultaneously understanding the visual-environment and fingertip-sensations. The visual encoder utilizes a  ViT\text{-B/16} \cite{CLIP}  to extract global features $\mathbf{z}_t^{v}$. To address tactile semantic ambiguity, we introduce a kinematics-conditioned encoder that takes the hand state $h_t$ as a prior input $\mathbf{z}_t^{h} = f_h(h_t)$, ensuring an automatic correspondence between features and hand poses. After tactile information is processed through a three-layer CNN to extract features $f_x(X_t)$, an MLP is used for non-linear fusion with the kinematics vector to generate a space-anchored representation:

$$\mathbf{z}_t^{x} = \phi(f_{\mathrm{x}}(X_t) \oplus \mathbf{z}_t^{h}) \eqno{(5)}$$

During pre-training, the visual encoder is frozen to serve as a semantic anchor. A contrastive loss $\ell(\mathbf{u}, \mathbf{v})$ is defined to align visual and tactile representations. The perception weights are trained by optimizing the objective function $\mathcal{L}_{\mathrm{pre}}$:

$$\mathcal{L}_{\mathrm{pre}} = \mathbb{E}_t [ \ell(\mathbf{z}_t^{v}, \mathbf{z}_t^{x}) + \lambda \ell(\mathbf{z}_t^{v}, \mathbf{z}_{t+1}^{x}) ] \eqno{(6)} $$
$$\ell(\mathbf{u}, \mathbf{v}) = -\log \frac{e^{\mathrm{sim}(\mathbf{u}, \mathbf{v})/\tau}}{\sum_{\mathbf{v}' \in \mathcal{N}} e^{\mathrm{sim}(\mathbf{u}, \mathbf{v}')/\tau}} \eqno{(7)}$$

where Eq.7 is the InfoNCE loss \cite{INFONCE}. The first term of $\mathcal{L}_{\mathrm{pre}}$ achieves visuo-tactile semantic alignment, while the second term utilizes temporal consistency constraints to predict tactile evolution using current visual information. This paradigm allows the model to effectively filter out noise and extract stable features, providing reliable perception for dexterous manipulation.

\textbf{Policy Learning via ACT.} In the policy learning stage, the pre-trained encoder is integrated into Action Chunking with Transformers (ACT) \cite{11aloha}. The network receives the joint observation sequence $e_t=[\mathbf{z}_t^{v}, \mathbf{z}_t^{x},\mathbf{z}_t^{h}]$ and predicts the action sequence $\hat{a}_{t:t+H-1}$. Leveraging ACT's one-time block generation mechanism, the system effectively mitigates control latency and jitter, improving the smoothness of human-robot collaboration. The model is optimized by minimizing the Mean Squared Error (MSE) loss.

\section{EXPERIMENTS}
We design experiments to address the following three core questions:

 \begin{itemize}
     \item \textbf{Q1. Policy Learning and Deployment:} By integrating visuo-tactile-kinesthetic multimodal feedback and the proposed method, can the collected data overcome the bottlenecks of vision-only policies and improve success rates in complex tasks?
     \item \textbf{Q2. Data Collection Efficiency and Scalability:} Compared to alternative solutions, can DexViTac achieve efficient collection comparable to natural human hands and support the construction of large-scale multimodal datasets?
     \item \textbf{Q3. Hand Tracking Robustness:} Can physical motion capture overcome the pose distortion inherent in vision-based tracking to provide high-fidelity expert demonstrations for policy learning?
 \end{itemize}

\subsection{Dataset and Experimental Design}

\begin{figure*}[htpb]
\centering
\includegraphics[width=0.9\linewidth]{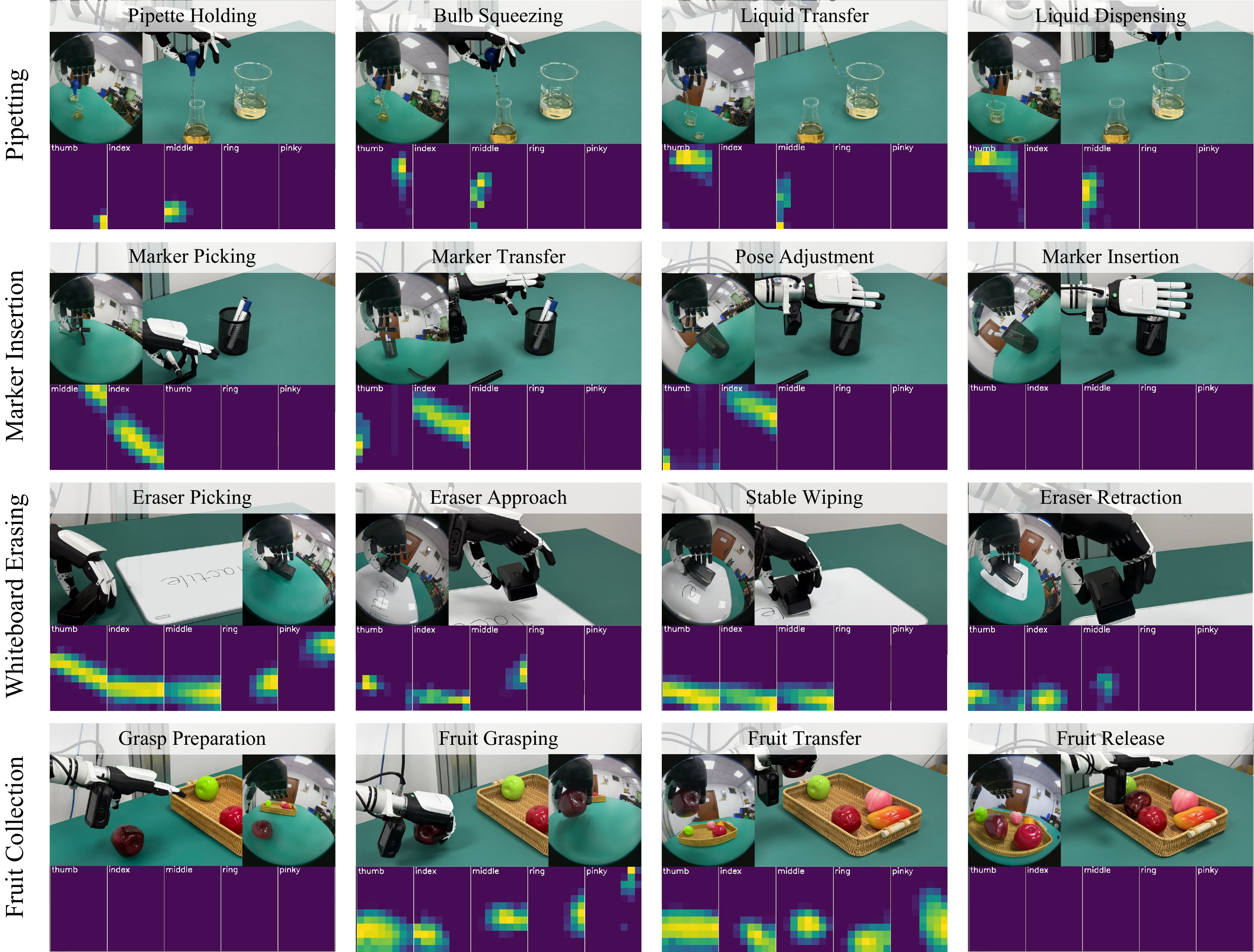}
\caption{\textbf{Real-world experimental deployment.} The figure illustrates the deployment of the proposed full method across four representative tasks: pipetting, whiteboard erasing, pen insertion, and fruit collection.}
\label{fig:traj}
\end{figure*}

\begin{figure}[h]
\centering
\includegraphics[width=1\linewidth]{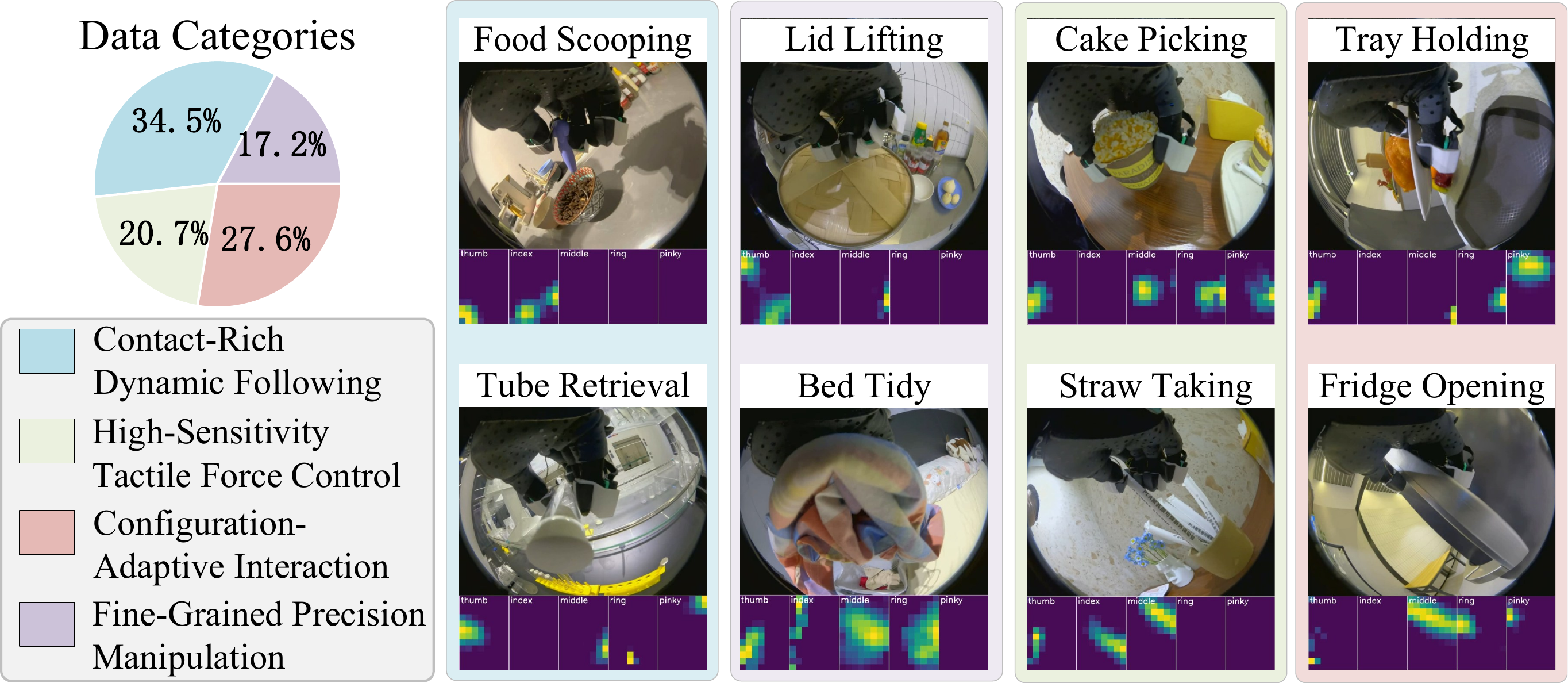}
\caption{\textbf{The DexViTac dataset.} The dataset contains 2,400$+$ visuo-tactile-kinematic demonstrations across 40$+$ tasks in 10$+$ real-world environments.}
\label{fig:dataset}
\end{figure}

As illustrated in Fig. 6, leveraging the high-efficiency collection capabilities of DexViTac, we have developed a large-scale dexterous manipulation dataset tailored for real-world, contact-rich scenarios. This dataset encompasses four task categories: High-Sensitivity Tactile Force Control, Contact-Rich Dynamic Following, Fine-Grained Precision Manipulation, and Configuration-Adaptive Interaction. Spanning more than 10 real-world environments(including kitchens, homes, retail spaces, and offices), the dataset comprises over 2,400$+$ visuo-tactile-kinematic demonstrations across 40+ daily tasks.

We select one representative task from each of the four categories to evaluate the multimodal dataset and the proposed policy strategy. The task selection prioritizes scenarios where single visual feedback or low-DOF grippers are insufficient, aiming to highlight the contribution of the multimodal data pipeline:\textbf{1) Pipetting:} The dexterous hand precisely regulates pinch force to operate a flexible rubber bulb, testing the high-sensitivity force control performance of the system. \textbf{2) Whiteboard Erasing:} This requires the end-effector to maintain constant contact pressure along dynamic trajectories, evaluating force compliance and stability during dynamic interactions. \textbf{3) Marker Insertion:} Grasping and inserting small markers, involving stable grasping and in-hand manipulation capabilities. \textbf{4) Fruit Collection:} Operating on heterogeneous objects with varying stiffness to verify adaptability to different physical properties. The experiments utilize a RealMan RM\text{-}65 robot arm and an Inspire RH56DFTP five-finger dexterous hand. Both the tactile sensors embedded in the dexterous hand and the human demonstration interface utilize the HIT Longlin-$96$ modules, maintaining sensor-level consistency.

\subsection{Task-Level Deployment and Generalization}

We compare our method with the following baseline methods:

\begin{itemize}
\item \textbf{Ours(Full Method):} Our complete proposed approach. It integrates kinematics-grounded and temporally consistent pre-trained encoders, fine-tuned with ACT. 
\item \textbf{B1.Vision-only ACT:} Trained using only visual observations, stripping away tactile feedback. 
\item \textbf{B2.Ours w/o Pretraining:} Removes self-supervised pre-training, employing end-to-end training with a randomly initialized encoder. 
\item \textbf{B3.Ours w/o Spatio-temporal Grounding:} Removes the kinematics prior ($h_t$) and the temporal consistency constraint ($\lambda=0$) during pre-training.
\end{itemize}
\begin{table}[h]
\centering
\caption{\centering Success Rates (\%) of Different Policies}
\vspace{6pt}
\renewcommand{\arraystretch}{1.2}
\setlength{\tabcolsep}{4pt}

\begin{tabularx}{\columnwidth}{cYYYYc}
\toprule[1.2pt]
\textbf{Method} 
& \textbf{Pipetting} 
& \textbf{Board Erasing} 
& \textbf{Marker Insertion} 
& \textbf{Fruits Collection}
& \textbf{Avg.} \\
\midrule
B1 & 6.7 & 16.7 & 13.3 & 33.3 & 17.5 \\
B2 & 43.3 & 56.7 & 40.0 & 53.3 & 48.3 \\
B3 & 70.0 & 46.7 & 36.7 & 80.0 & 58.4 \\
\textbf{Ours (Full)} 
& \textbf{83.3} 
& \textbf{86.7} 
& \textbf{80.0} 
& \textbf{93.3} 
& \textbf{85.8} \\
\bottomrule[1.2pt]
\end{tabularx}

\vspace{2pt}
\hfill{\footnotesize Each task is evaluated over 30 trials.}

\end{table}

As shown in Table I, the full method performs best across all tasks. The ablation analysis is detailed below:

\textbf{Efficacy of Perceptual Pre-training (B2 vs. Ours):} In pipetting and fruit collection tasks, the success rate of the baseline without pre-training (\textbf{B2}) drops to 43.3\% and 53.3\%, respectively. These results are significantly lower than the 83.3\% and 93.3\% achieved by the full method, highlighting the necessity of the proposed tactile-kinematic representations. This outcome demonstrates that self-supervised pre-training, through the construction of robust representations, effectively alleviates the convergence difficulties associated with multimodal high-dimensional features. Consequently, this enables the policy to accurately infer object deformation and topological characteristics from the tactile data stream.

\textbf{Necessity of Spatio-Temporal(ST) Coupling (B3 vs. Ours):} Removing the ST coupling (\textbf{B3}) caused the success rates for the marker insertion and whiteboard erasing tasks to drop to 36.7\% and 46.7\%, respectively. \textbf{i) Spatial aspect:}  In the marker insertion task involving in-hand adjustment, the lack of kinematic priors prevents the system from resolving tactile semantic ambiguity. \textbf{ii) Temporal aspect:}  In the erasing task, removing temporal consistency ($\lambda$) causes significant pressure fluctuations. This validates that our predictive objective effectively models contact dynamics, ensuring smooth closed-loop force control.

\textbf{Essential Need for Multimodal Closed-loop (B1 vs. Ours):} In tasks characterized by visual occlusion (pipetting) or a heavy reliance on force feedback (erasing), the vision-only scheme (\textbf{B1}) yields success rates of only 6.7\% and 16.7\%. This highlights that the tactile modality is irreplaceable for compensating for visual loss and achieving high-precision force-controlled manipulation.

\subsection{Data Collection Efficiency}

\begin{figure}[!h]
\centering
\includegraphics[width=0.9\linewidth]{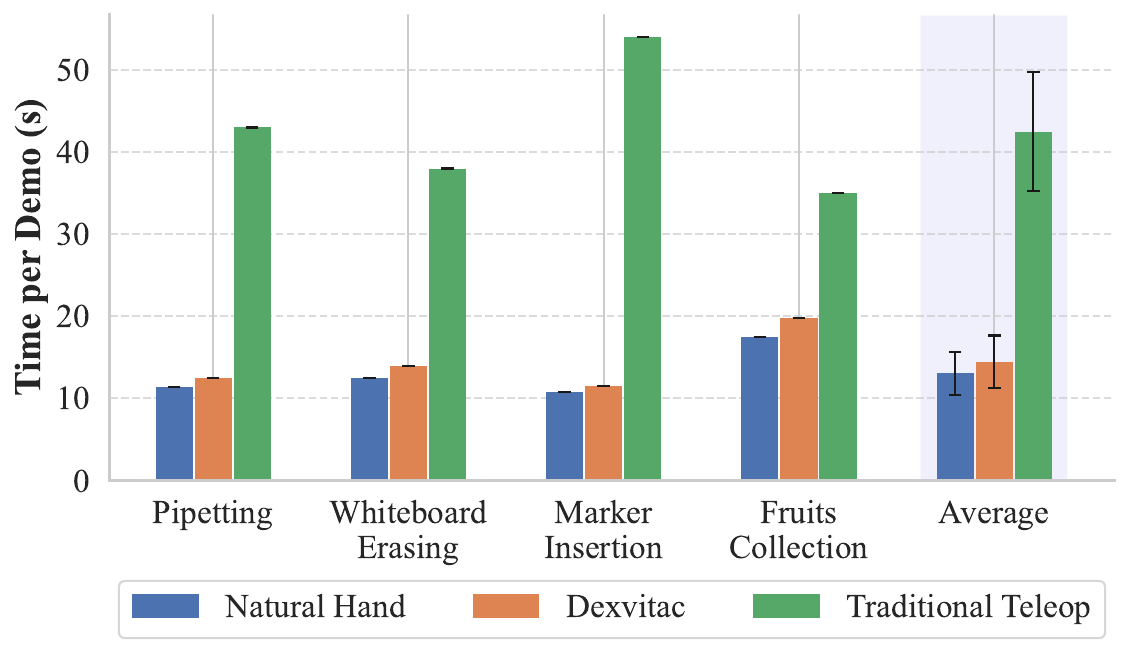}
\caption{\textbf{Data collection efficiency.} Compared with traditional teleoperation, DexViTac significantly accelerates the acquisition of demonstration trajectories, achieving an effective throughput of over 248 demonstrations per hour, which closely approaches natural human operational speeds.}
\label{fig:efficiency}
\end{figure}

As illustrated in Fig. 7, the average collection durations for DexViTac across the four tasks are 12.5 \text{s}, 14.0 \text{s}, 11.5 \text{s}, and 19.8 \text{s}, respectively, with an overall mean of 14.5 \text{s/demo}. In contrast, natural human demonstration requires 13.1 \text{s/demo}, while traditional teleoperation takes 112.3 \text{s/demo}. Consequently, the data collection efficiency of our system exceeds 248 \text{demos/hour}, which is significantly higher than the 84 \text{ demos/hour} of teleoperation and approaches the 275 \text{demos/hour} efficiency of natural human hands. As task complexity increases, the efficiency advantage of our system becomes even more pronounced, providing robust support for constructing large-scale datasets and learning generalized policies in real-world environments.

\subsection{Hand Pose Tracking Quality}

Existing dexterous manipulation data collection systems often rely on vision-based models to track hand poses\cite{Dexmv}. However, in contact-intensive tasks, hand-object occlusions and self-occlusions frequently cause vision algorithms to fail, which negatively affects action execution. To verify the reliability of DexViTac data during complex physical interactions, we compare the data quality of our system with RTMPose \cite{DBLP:journals/corr/abs-2303-07399}, a representative vision-based tracking algorithm.

In fruit grasping, the vision-based baseline (RTMPose) exhibits high-frequency jitter and step-like jumps under hand-object occlusion. Conversely, DexViTac produces smooth, continuous joint trajectories, demonstrating superior robustness to visual interference.

\begin{figure}[h]
\centering
\includegraphics[width=0.8\linewidth]{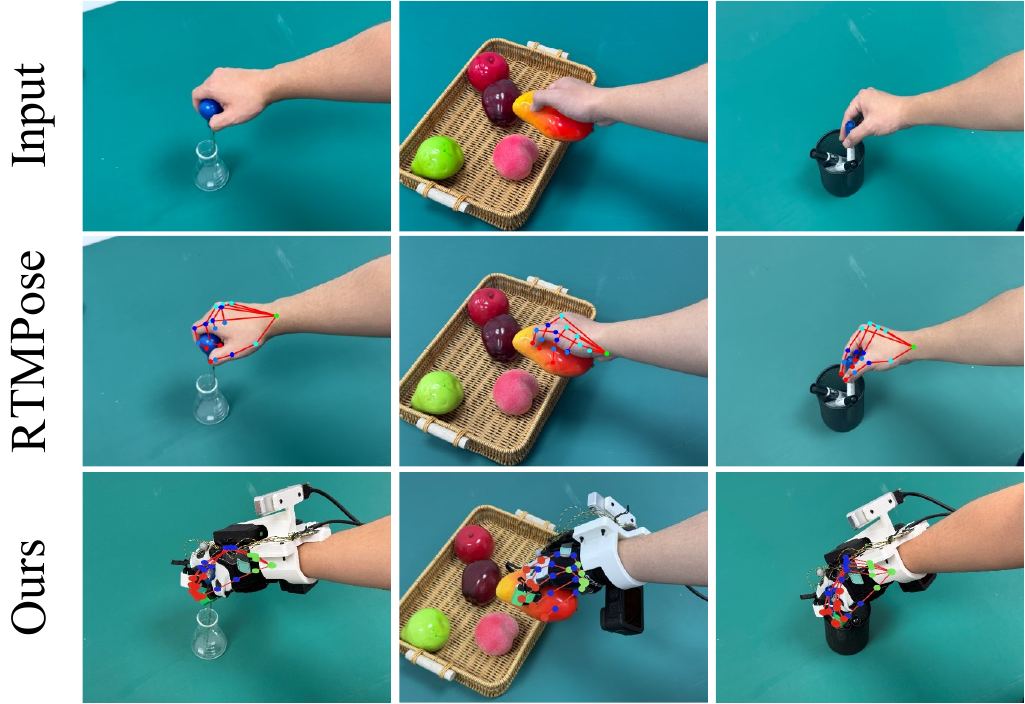}
\caption{\textbf{Hand pose tracking under occlusion.} Vision-based methods degrade sharply under severe hand--object occlusions, whereas our physical method remains robust and consistently provides high-fidelity, fine-grained hand articulation.}
\label{fig:vision}
\end{figure}

As illustrated in Fig. 8, in tasks such as pipetting, fruit collection, and marker insertion, the vision-only approach often suffers from pose estimation distortion and physical interpenetration due to severe self-occlusion and hand-object mutual occlusion. In contrast, DexViTac, relying on physical motion-capture gloves, consistently outputs high-fidelity hand poses that adhere to real physical constraints.

\begin{figure}[h]
\centering
\includegraphics[width=0.85\linewidth]{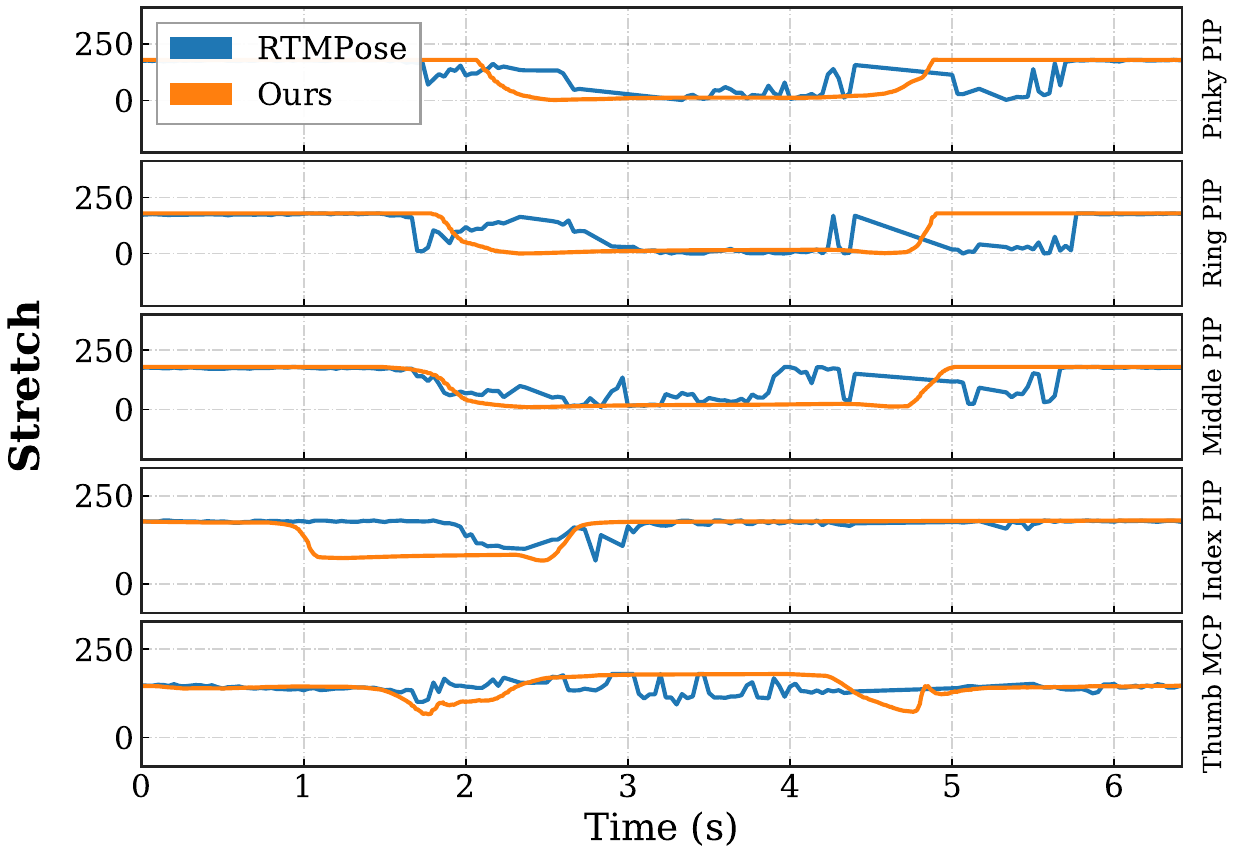}
\caption{\textbf{Stability of hand pose Tracking.} DexViTac provides smooth and continuous trajectories during grasping, whereas vision-based baselines exhibit significant high-frequency noise and sudden jumps.}
\label{fig:occlusion}
\end{figure}

Taking the hand-object mutual occlusion during fruit grasping as an example, we quantitatively analyzed the tracking curves of representative joints across the five fingers (Fig. 9). It is observed that between 1.5 \text{s} and 5 \text{s}, the vision-based baseline RTMPose exhibits significant high-frequency jitter and sudden step-like mutations. In contrast, our solution demonstrates complete robustness against visual occlusion, maintaining smooth and continuous trajectories throughout. Such low-noise, coherent kinematic data is essential for policy learning, which fundamentally proves the necessity of the DexViTac hardware design.

\section{CONCLUSION}
This paper presents DexViTac, a human-centric multimodal data collection for contact-rich dexterous manipulation. At the hardware level, DexViTac achieves native synchronous acquisition of multimodal data, including high-resolution tactile arrays and 19\text{-DOF} hand kinematics. At the algorithmic level, it effectively resolves semantic ambiguity in multi-finger contact through kinematics-anchored representation learning. Experimental results demonstrate that the system achieves a large-scale collection efficiency exceeding 100\text{ demos/hour}, and the learned policy attains an average success rate of 85\% in representative tasks, significantly outperforming the baselines.

\textbf{Limitation \& Future Works:} While DexViTac offers significant advantages in perception and the scale of data collection, the fixed-base setup limits the reach for long-horizon operations, and the single-arm modality remains insufficient for complex bimanual asymmetric coordination tasks. In future, we aim to develop a dual-arm architecture to facilitate high-dimensional data collection and incorporate mobile platforms to overcome spatial constraints. Ultimately, this will enable the study of whole-body dexterous manipulation in challenging field environments.

\section*{ACKNOWLEDGMENT}

This work was supported in part by the Fundamental and Interdisciplinary Disciplines Breakthrough Plan of the Ministry of Education of China under Grant JYB2025XDXM208, the National Natural Science Foundation of China under Grant 52575019 and 52188102, and the National Natural Science Foundation of China
Overseas Excellent Young Scholars Fund.

\bibliographystyle{IEEEtran}
\bibliography{refs}

\end{document}